\documentclass[runningheads]{llncs}
\usepackage[T1]{fontenc}
\usepackage{graphicx}
\usepackage{arydshln}

\usepackage{booktabs}
\usepackage{tabularx}
\newcommand{\minus}{\scalebox{0.8}{$-$}}
\newcommand{\plus}{\scalebox{0.8}{$+$}}
\RequirePackage{amsmath}
\usepackage{amsfonts}
\usepackage{amsmath}
\usepackage[dvipsnames]{xcolor}

\usepackage{url}
\newcolumntype{?}{!{\vrule width 1.1pt}}

\usepackage[numbers]{natbib}

\usepackage[colorlinks=true, linkcolor=blue, urlcolor=blue, citecolor = blue]{hyperref}

\usepackage{parskip}
\begin{document}
%

\title{\texttt{Konstruktor}: A Strong Baseline for  \\ Simple Knowledge Graph Question Answering} 

\titlerunning{Konstruktor: A Strong Baseline for  Simple KGQA}

\author{Maria Lysyuk\inst{1,2} \and
Mikhail Salnikov\inst{1,2}\and
Pavel Braslavski\inst{3}  \and \\
Alexander Panchenko\inst{1,2}}

\authorrunning{Maria Lysyuk, Mikhail Salnikov, Pavel Braslavski, and Alexander Panchenko}

\institute{Skolkovo Institute of Science and Technology, Russia \and
Artificial Intelligence Research Institute, Russia \and
Nazarbayev University, Kazakhstan \\ \email{ \href{mailto:maria.lysyuk@skol.tech}{maria.lysyuk@skol.tech}}}

\maketitle  
\begin{abstract}
While being one of the most popular question types, simple questions such as ``Who is the author of Cinderella?'', are still not completely solved. 
Surprisingly, even most powerful modern Large Language Models are  prone to errors when dealing with such questions, especially when dealing with rare entities. 
At the same time, as an answer may be  
one hop away from the question entity, one can try to develop a method that uses structured knowledge graphs (KGs) to answer such questions. 
In this paper, we introduce \texttt{Konstruktor} -- an efficient and robust approach that breaks down the problem into three steps: (i)~\textit{entity extraction and entity linking}, (ii)~\textit{relation prediction}, and (iii)~\textit{querying} the knowledge graph.
Our approach integrates language models and knowledge graphs, exploiting the power of the former and the interpretability of the latter.
We experiment with two named entity recognition and entity linking methods and several relation detection techniques.
We show that for relation detection, the most challenging step of the workflow, a combination of relation classification/generation and ranking outperforms other methods.
We report \texttt{Konstruktor}'s strong results  on four datasets.

\end{abstract}

\section{Introduction}





Large open Knowledge Graphs (KGs), such as Wikidata, store myriads of facts about the real world in the form of \textsf{<subject, relation, object>} triples, e.g. \textsf{<Albert\_Einstein, born\_in, Ulm>}. These are often complementary to knowledge expressed in texts, for instance, people rarely write explicitly about basic facts, common knowledge, or about rare events, especially in narrow domains. For such reasons, KGs are proven to be valuable resources for various NLP tasks, including question answering~(QA). Although even the largest KGs have limited coverage compared to information in the vast collections of unstructured text documents available online, the structure of KGs makes the information they contain easier to verify and manipulate. These features make KGs a suitable resource, for example, for testing the veracity of responses returned by the Large Language Models~(LLMs)~\cite{pan2023unifying}. 

In this work, we address the task of answering simple questions over KGs. In this case, the answer entities are one hop away from the entities mentioned in the question in the KG. Despite their name, end-to-end LLM solutions without direct access to KGs often fail to answer simple questions \cite{razzhigaev2023system}.
In our study, the answer to the question is looked up directly in the KG, which allows us to get the up-to-date and related answer. Moreover, recent studies~\cite{sun2023head} show that LLMs are good only for answering questions about popular entities, which is further supported and analyzed in our study. 
In contrast, our method is much less sensitive to the popularity of the entities.


The term ``simple questions'' was coined by Bordes et al.~\cite{bordes2015large}, who also released a dataset with the same name. 
The question mentions the KG triple's subject and relation, with the object being the expected answer. 
For instance, the question ``Who is the author of Cinderella?'' corresponds to the KG triple \textsf{<Cinderella, author, Charles Perrault>}. 
Given the correctly identified subject ``Cinderella'' and the relation ``author'', the object will be returned as the answer of interest. 
Original SimpleQuestions dataset uses Freebase~\cite{bollacker2008freebase} as the underlying KG. Freebase was discontinued in 2016, and the data was imported to Wikidata~\cite{pellissier2016freebase}. 
New KGQA benchmarks are being created on the basis of Wikidata and existing ones are being migrated to it, which can be vividly seen in a curated KGQA leaderboard.\cite{perevalov2022knowledge}.


Mohammed et al.~\cite{DBLP:conf/naacl/MohammedSL18} raise a question whether  complex neural architectures are really needed to answer simple questions. 
Authors show that the actual improvements are much more modest than the literature suggests. Although the parts of the \texttt{Konstruktor} use pre-trained language models, we follow the same principle of creating a simple yet effective approach.

In our work, we revisit a classical approach to KGQA \cite{DBLP:conf/naacl/MohammedSL18,hu2023empirical}, motivated by the structure of the triples stored in the KG and the structure of the corresponding SPARQL query that parses it to obtain the answer. The proposed approach,  \texttt{Konstruktor} consists of three
components: (i) detection of the subject in the question and its linking to the entity from the KG; (ii) linking the relation in the question to the KG scheme; (iii) generating and executing a corresponding SPARQL query. In addition, we implement several variations of each component and provide practical recommendations for choosing a specific configuration.

The advantages of this approach include the use of off-the-shelf named entity recognition (NER) and entity linking (EL) components, and much more data- and 
compute-efficient training procedures compared to end-to-end or LLM-based approaches. In this study, we have developed and compared a number of relation detection models. In contrast to previous studies, we focus on datasets with Wikidata annotations and use the Wikidata endpoint\footnote{\url{https://query.wikidata.org/sparql}} to run SPARQL queries. This provides a more realistic scenario, for reproducibility of the experiments we downloaded the respective Wikidata dump actual for the time of conducting experiments\footnote{\url{https://sc.link/Dyely}
}.
Our study follows the work by Mohammed et al.~\cite{DBLP:conf/naacl/MohammedSL18}, however, we experiment with next-generation tools, the Wikidata KG, and four datasets.

The contributions of our study are as follows:
\begin{enumerate}
    \item We present \texttt{Konstruktor}, a component-based baseline for simple KGQA  competing and in some cases outperforming various baselines including computationally expensive end-to-end neural methods;
    \item A comparison of the performance of the SOTA entity linkers and relation detection in the context of question answering;
    \item An open implementation of the system.\footnote{
The code is available at
\url{https://github.com/s-nlp/konstruktor}}
\end{enumerate}

\section{Related Work}

\vspace{-1ex}

Traditional KGQA methods can be divided into two types: retrieval-based and semantic parsing. Semantic parsing approaches follow a parse-then-execute  para-digm: a question is parsed into a logical form, such as SPARQL, and executed against the KG to obtain the answer. The closest to the current approach is the paper by Lukovnikov et al.~\cite{lukovnikov2019pretrained}. However, the authors use an inverted index to link ID to the entities, thus not considering the entity disambiguation step. Next, only relation classification is used for relation detection, thus limiting the predictive power to the relations seen in the training data. Other popular approaches are Falcon 2.0 \cite{sakor2020falcon} and QAnswer \cite{diefenbach2020towards}. Falcon 2.0 \cite{sakor2020falcon} relies on fundamental principles of English morphology and makes joint entity and relation linking using n-gram tiling and n-gram splitting. QAnswer \cite{diefenbach2020towards} uses all candidates that are intersected with the n-grams from the questions except stopping words, makes a breadth-first search of depth 2 starting from the found candidates to collect triples, and generates candidate SPARQL queries that are further ranked.

In contrast, retrieval-based techniques involve the transformation of textual questions into vectors, which are then mapped onto a vector space containing potential entities and predicates. One of the pioneering approaches in this area is the Knowledge Embedding based Question Answering (KEQA) framework \cite{Huang2019KnowledgeGE}. By separately training embeddings for the entity and the relation from the question respectively, and building an entity extraction module, the authors use a carefully designed joint distance metric, where the closest fact of the three learned vectors in the KG is returned as the answer. In more recent approaches, such as M3M~\cite{razzhigaev2023system}, the authors use separate projections of the question text embeddings to the subject, predicate and object of the KG triple, and compute the distance between these learned embeddings and the pre-trained graph embeddings of the KG triples.

Recently, a third wave of approaches has emerged based on the pre-trained LLMs such as T5~\cite{DBLP:journals/jmlr/RaffelSRLNMZLL20} or GPT~\cite{radford2018improving,radford2019language}. To provide a comparison with this type of approach, we conducted experiments with FlanT5-xl\cite{chung2022scaling} and t5-xl-ssm-nq\cite{roberts-etal-2020-much}, which have been shown to be effective for question answering, as demonstrated by Roberts et al. \cite{roberts-etal-2020-much}, GPT-3 \cite{brown2020language}, which shows impressive performance in a wide range of benchmarks, and ChatGPT \cite{openai2023gpt} as one of the current leading systems in the field of Natural Language Processing.

\vspace{-2ex}

\section{Datasets}
\label{data}

In our experiments, we use four datasets linked to Wikidata: SimpleQuestionsWD, RuBQ-en, Mintaka, and LC-QuAD 2.0. Table~\ref{dataset_stats_main} summarises the statistics of the datasets. For SimpleQuestionsWD and RuBQ-en, the subsets of 1-hop questions were taken to align our results with other existing approaches (see Section~\ref{res_section} for details).\footnote{
The data is available at
\url{https://github.com/s-nlp/konstruktor}}


\begin{table}[h!]
\small
\centering
\renewcommand{\arraystretch}{1.1}
\caption{Statistics of the datasets in our experiments. \#Relations -- 
the number of unique Wikidata relations in the dataset; * denotes subset of simple (1-hop) questions.}

{
\setlength\tabcolsep{0.35cm}
\begin{tabular}{lrrrr}

\toprule

\bf Dataset             & \textbf{\#Relations} & \textbf{Train} & \textbf{Dev} & \textbf{Test} \\ \midrule
SimpleQuestionsWD~\cite{SQ_WD}          & 117   & 19,481    & 2,821     & 2,491 \\
RuBQv2.0-en*~\cite{rybin2021rubq}           & 117   &--         & 373       & 1,186 \\
LC-QuAD 2.0*~\cite{dubey2019lc}          &  794    & --     & --     & 824\\
Mintaka* ~\cite{sen2022mintaka}         &   --    & 8,289     & 1,199     & 173\\
SMART RL 2022*~\cite{mihindukulasooriya2021semantic} & 3,095 &~~24,112   & --        & -- \\ 
\bottomrule
\end{tabular}
}
\vspace{-15pt}
\label{dataset_stats_main}
\end{table}

\begin{enumerate}

\item \textbf{SimpleQuestionsWD} is a subset of the SimpleQuestions \cite{bordes2015large} (further SQ-WD) linked to Wikidata~\cite{SQ_WD}. The original SQ dataset~\cite{bordes2015large}  consists of 100k+ questions that were generated by human annotators based on presented Freebase triples. 
Since the questions were generated based on the sampled Freebase triples, it can lead to a situation, when a correct SPARQL query corresponding to the original triple would return multiple answers, only one of which is correct according to the dataset. For example, for the question ``What male actor was born in Warsaw'', the only correct answer is ``Michal Grudziński'', while there are obviously a number of correct answers. This makes it impossible to solve the SQ-WD completely: the upper bound of the accuracy due to the ambiguity of the answers is estimated to be 83.4\% \cite{petrochuk2018simplequestions}. 

\vspace{1.1ex}

\item \textbf{RuBQ v2.0} is a relatively small Russian KGQA dataset comprising questions from two sources: online quizzes (they constitute version 1, see \cite{korablinov2020rubq})  and search query suggestion APIs~\cite{rybin2021rubq}. Due to its modest size, it has only official development and test splits and no training part. 
We use a subset of machine-translated English 1-hop questions from the official dataset. 



\vspace{1.1ex}

\item \textbf{Mintaka} dataset \cite{sen2022mintaka}, similar to RuBQ, has more natural questions than SQ-WD. It contains spans for the subjects that make NER training easier. The main drawback of this dataset is the lack of annotated relations. To obtain the subset of 1-hop questions, we first filtered the questions where the answer is one hop away from the entity in the question to obtain the set of relations between subject and object. Then, among these relations, the one leading to only one answer was selected.

\vspace{1.1ex}

\item \textbf{LC-QuAD 2.0} dataset \cite{dubey2019lc}, is a large dataset of 30,000 complex questions with corresponding SPARQL queries for Wikidata. The subset of 1-hop questions for this paper was obtained by selecting all questions with corresponding "simple" SPARQL queries, where the question is addressed to either object or subject. 

\end{enumerate}

\vspace{1.1ex}

As an additional dataset for the relation detection part we use:

\vspace{1.1ex}

\textbf{SMART RL 2022}~\cite{mihindukulasooriya2021semantic} is a dataset associated with the relation linking task at open challenge SMART 2022.\footnote{\url{https://smart-task.github.io/2022}} The questions in the dataset are gleaned from a variety of KGQA datasets.
As we focus on simple questions, we filtered out questions with more than one property. The dataset is only used for relation-linking training.

\section{\texttt{Konstruktor}: a Component-based KGQA Method}

\begin{figure*}[ht!]
\begin{center}
\includegraphics[width=\textwidth]{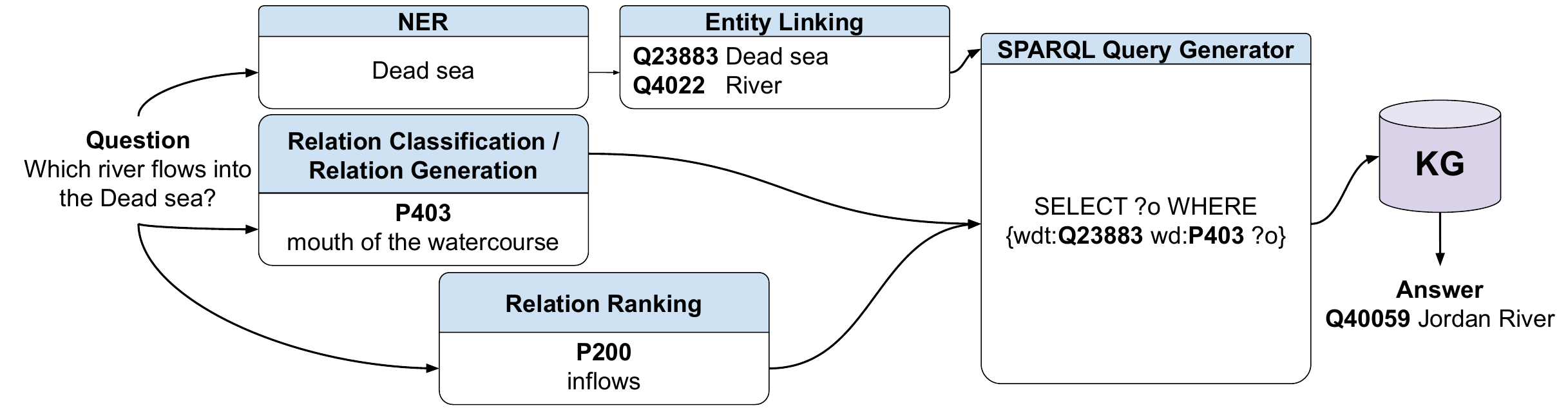} 
\caption{An overview of the \textsf{Konstruktor} method workflow for simple question answering.}
\label{main_pipe}
\end{center}
\end{figure*}

The present study introduces a new method \texttt{Konstruktor}, for addressing simple questions, which is based on the principles of the Semantic Web. The proposed approach implements a set of techniques for handling entities and generating SPARQL queries. 
Our workflow includes three stages: 1)~entity detection and linking, 2)~property detection, and 2)~SPARQL query formation to obtain the final answer. The workflow is illustrated in more detail in the diagram in Figure~\ref{main_pipe}.



\subsection{Entity Extraction and Entity Linking}

In \texttt{Konstruktor}, we consider two cutting-edge entity linking models: the current SOTA multilingual entity disambiguation model mGENRE \cite{de2020autoregressive} and the recently developed BELA model \cite{plekhanov2023multilingual}.


The mGENRE is a sequence-to-sequence transformer-based model predicting the Wikidata Entity ID for the input sentence. This model expects the named entities to be highlighted in a text that forced us to develop the Named Entity Recognition~(NER) model. We tried to use available NER models such as Spacy, SparkNLP, and Stanza (selected as the top ones according to the recent review of NER systems~\cite{vajjala2022we}), but they did not show satisfactory results in our experiments.  On the SQ-WD dataset, Stanza, Spacy, and SparkNLP obtained 74\%, 64\%, and 88\% error rates correspondingly. Since Spacy model showed the best performance, 
it was chosen as the default configuration for further fine-tuning.\footnote{\url{https://spacy.io/usage/training}} One of the reasons for low performance of pre-trained models is the high percentage of low-case named entities in the SQ-WD (for the detailed analysis refer to Section~\ref{ent_lin_discussion} and Table \ref{low_case}). 


BELA~\cite{plekhanov2023multilingual} is an end-to-end multilingual entity linking model.  It is composed of a Mention Detection, Entity Disambiguation, and Rejection Head. It belongs to the class of the dense-retrieval-based models: at the Entity Disambiguation stage a bi-encoder architecture is used and a kNN search between the mentioned encoding and candidate entity encodings is realised.
Although in the original paper, BELA falls short of the mGENRE model, it doesn't require a separate NER module (mGENRE is just an entity disambiguation module and works badly without highlighting the named entities in a text). In addition, the mGENRE model is more computationally
expensive at the inference stage.

As a result of the entity linking step, both the entity IDs from the question and the corresponding model's confidence score in the given prediction $score(s_i)$ are returned, which will be further used in the relation ranking step. 

\subsection{Relation Detection}

As a bottleneck for the overall performance of KBQA systems, relation linking has proven to be a challenging task, with SOTA techniques achieving less than 50\% F1-score on a number of datasets \cite{lin2020kbpearl,mihindukulasooriya2020leveraging,sakor2019old}. Similar to us, more recent approaches such as GenRL~\cite{rossiello2021generative} show great relation detection results on LC-QuAD 2.0 \cite{dubey2019lc} and SQ-WD~\cite{SQ_WD} using generative relation linking approach, but there is no open code implementation for a training procedure. Two main challenges are: (i)~training data is often limited; (ii)~relations in text and KG are often lexicalized differently (implicit mentions).

We propose three methods for relation prediction: relation classification, relation ranking,  and relation generation. Depending on the method, the final SPARQL queries are formed differently. For relation classification or relation generation, given the top-n predicted subjects and the top-1 relation, $n$ SPARQL queries are formed. They are iterated until the first non-empty answer is found. The iteration procedure is necessary because a relation classifier/generator may return relations that do not belong to the top-1 predicted subject. The ranking approaches return only one entity and one relation by design, resulting in a single SPARQL query.

All three methods of relation detection have their advantages and disadvantages. In most cases, relation classification or relation generation has a better accuracy than the other two approaches.  However, the procedure does not guarantee that any answer will be found. 
Given the limited predictive power of the relation classification and generation (it is impossible to predict relations not seen in the training data or to link text-generated labels to the list of predefined labels from the KG respectively), ranking can generate correct labels if relation classification and generation fail. This motivates a joint procedure of the SPARQL query generation in the proposed \texttt{Konstruktor}: given the higher accuracy, queries with relations predicted by the classifier or the generator are tried first.  If all of them return empty answers, the SPARQL query using ranking  guarantees a non-empty answer. 


\vspace{1.1ex}

\textbf{Classification}
%
problem was successfully solved by using the transfer learning approach, which is based on the pre-trained BERT-base-uncased model~\cite{devlin2018bert} with Binary Cross Entropy (BCE) loss. Furthermore, following the paper by ~\cite{huang2021balancing}, the method has been improved using distribution-balanced (DB) loss. This was necessary because the standard BCE loss is sensitive to class imbalance, 
as pointed out by the authors. This loss takes into account rebalanced weighting, which leads to a reduction of redundant information in the label co-occurrence and the problem of negative tolerant regularisation.  In many QA datasets, there are many examples of just one question per relation in the training data, thus insertion of such a loss 
can dramatically increase the overall performance.



\vspace{1.1ex}

\textbf{Ranking} follows the pipeline proposed by Hu et al.~\cite{hu2023empirical}.
First, the entity in a question is replaced by the token \textsf{<e>} to obtain the question pattern \textit{p}. For example, \textit{p} for the question ``What did Jane Austen write?'' is ``What did \textsf{<e>} write?''.
Second, the correct relation $r^{\plus}$ is complemented with $k$ negative $r^{\minus}$ relations randomly sampled from the rest of question entity's relations. 

Given each relation $r_i \in \mathbb{R}_C$, where $\mathbb{R}_C$ is the set of relations of the given subject, the predicted relation $\hat{r}$ is defined as
\begin{equation}
    \hat{r} = \underset{r_i \in \mathbb{R}_C}{argmax} \hspace{0.5ex} Score (p, r_i) 
\end{equation}

\begin{equation}
Score(p, r_i) = cos(Pool(h_p), Pool(h_{r_i})),
\end{equation}
 where $h_p$ and $h_{r_i}$ are embeddings obtained by pretrained LM and $Pool()$ is a pooling layer. To maximise the margin between gold $r^{\plus}$ and  negative $r^{\minus}$ relations authors use hinge loss during training:

\begin{equation}
\mathcal{L} = \sum_{i=1}^k \{0, \gamma - Score(r^{\plus})+Score(r_i^{\minus})\},
\end{equation}

where $\gamma$ is a constant.


Given the list of ranked relations with their scores $score(r_i)$ and the ranked subjects with their scores $score(s_i)$ (in our case these scores are either scores of mGENRE or BELA models), the best pair for the SPARQL query after this step is defined as the one with the highest mean score $0.5*(score(r_i)+score(s_i))$.

\vspace{1.1ex}

\textbf{Relation generation} has been solved as a sequence-to-sequence problem with the T5-large~\cite{DBLP:journals/jmlr/RaffelSRLNMZLL20} model. We fine-tune the model to predict the labels of the corresponding properties. This simple approach has 
shown good results, but suffers from the fundamental problems described above: for instance, the possibility to generate only labels similar to the ones seen in train . The sequence-to-sequence model generates labels of relations that we need to link 
by matching the generated labels. This procedure requires preparing a mapping from labels to property IDs. On top of the labels of properties, we added all possible aliases of properties to this mapping. 



\subsection{SPARQL Query Generation}
\label{sparql_queries}

Generally, simple questions are the ones where given found subject and relation, object is the answer of interest. But in some cases, subject is the answer of interest. In other words, the place of the found entity in the question is crucial for further correct SPARQL query creation. Consider the following two questions:

\begin{enumerate}
    \item \textit{Where did roger marquis die?} This question has a subject \textsf{Roger Marquis} (Q7358590) and relation \textsf{place of death} (P20). The respective SPARQL query: \vspace{0.01cm}
    
    \textsf{SELECT ?o WHERE \{wd:Q7358590 wdt:P20 ?o \} \vspace{0.3cm}}  
    
    \item \textit{Which home is an example of italianate architecture?} This question has a object \textsf{italianate architecture} (Q615196) and relation \textsf{architectural style} (P149). The respective SPARQL query  \vspace{0.3cm}:
    
    \textsf{SELECT ?s WHERE \{?s wdt:P149 wd:Q615196\}\vspace{0.3cm}}
\end{enumerate}


In the SQ-WD dataset, such cases are specially labelled. Namely, for the first question, the relation has the label \underline{P}20, while for the second question, the relation has the label \underline{R}149 (meaning ``reverse''). Such labelling allows us to define the place of the entity found in the query for the classification task. 



\section{Experiments}


%
\label{res_section}
\vspace{-1ex}

\paragraph{Discussion.} In Table~\ref{tab:results} we have compared our proposed approach \texttt{Konstruktor} with other baselines. As described in the related work section, there are three main types of approaches: retrieval-based, semantic parsing, and sequence-to-sequence approaches. The first two rows (QAnswer~\cite{diefenbach2020towards} and FALCON 2.0 \cite{sakor2020falcon}) represent semantic parsing approaches, the next two rows are for the retrieval-based approaches (KEQA \cite{Huang2019KnowledgeGE} and M3M~\cite{razzhigaev2023system}), and the last five rows correspond to the sequence-to-sequence methods. In order to align our results with the M3M model, and thus with other baselines, we select the same subsets of the datasets used in the original paper~\cite{razzhigaev2023system}. Following the authors' logic, we successfully obtained the same test subsets for the RuBQ 2.0 (1186 questions). For the SQ-WD dataset, we did our best to reproduce the same subset (keeping only questions whose triples are present in Wikidata8M~\cite{korablinov2020rubq}), however, we obtained 2491 instead of 2438 questions. We hope that this minor discrepancy doesn't influence the overall comparison. 

\begin{table*}[ht!]
\small
\centering
\renewcommand{\arraystretch}{1.1}

\caption{Comparison of our system for the SQ-WD and RuBQ 2.0 datasets with KGQA baselines in terms of Accuracy@1 for one-hop QA datasets. The best scores are highlighted. *Subset of 1-hop questions.}
\scalebox{0.85}{
\begin{tabular}{lrrrr}
\toprule
\bf Model & \bf  SQ-WD & \bf  RuBQ-en* & \bf  LC-QuAD 2.0* & 
 \bf Mintaka*  \\ 
\toprule
\# test questions & 2,491 & 1,186 & 824 & 173\\
\midrule
QAnswer~\cite{diefenbach2020towards}  & 33.31 &  32.30 & - & - \\
FALCON 2.0 \cite{sakor2020falcon} &  24.89 &   13.74 & 29.73 & 16.76\\
KEQA \cite{Huang2019KnowledgeGE} --  PTBG  & 48.89 &  33.80 & -  & - \\ 
M3M~\cite{razzhigaev2023system} &  53.50   & 49.50  & 2.18 & 16.18\\
ChatGPT -- GPT-3.5-turbo & 17.75 &  30.12 & 9.71 & 72.25\\
GPT-3 -- davinci-003  & 28.51 &  34.20 & 9.83 & \textbf{80.35}\\
t5-xl-ssm-nq, zero-shot \cite{roberts-etal-2020-much} & 18.83 & 32.80 & 9.34 & 59.53\\
t5-xl-ssm-nq, fine-tuned \cite{roberts-etal-2020-much} & 30.68 & 32.80 & 12.80 &  60.69\\
FlanT5-xl, zero-shot~\cite{chung2024scaling} & 11.70 & 4.40 & 8.30 & 31.25\\
FlanT5-xl, fine-tuned~\cite{chung2024scaling} & 28.17 &  23.03 & 8.65 & 41.62\\

\midrule
\textsf{Konstruktor}: relation classification  & 58.33          & 46.30 & 18.08 & 53.76  \\
\textsf{Konstruktor}: ranking  & 52.99          & 45.28  & \textbf{59.10} & 58.38       \\
\textsf{Konstruktor}: label generation  & 56.16          & 45.36  & 44.17 & 58.96    \\
\textsf{Konstruktor}: relation classification + ranking  & \textbf{59.13} & 51.10 & 44.90 & 65.90 \\ 
\textsf{Konstruktor}: relation generation + ranking  & 57.05         & \textbf{52.45} & 55.34 & \textbf{67.63}  \\
\bottomrule
\end{tabular}
}

\vspace{-15pt}

\label{tab:results}
\end{table*}

As for our approach, in Table~\ref{tab:results} the results for the best combinations of NER and entity linkers obtained on the validation are presented (the details can be found in Section~\ref{Ablation}).
The final accuracy scores are shown depending on the relation detection model for the respective datasets on the test set.

The first three rows in the table are calculated for the solo implementation of each of the relation detection methods (relation classification, label generation, and ranking). The last two rows are the ensembles of two approaches. For instance, the last row means that if the relation generation method failed (no answers were found for all the SPARQL queries obtained by this method), the answer is obtained using the relation ranking approach.

Two ensembles, relation classification+ranking and relation generation+ranking, beat SOTA for the SQ-WD and RuBQ datasets. 
Interestingly, LC-QuAD 2.0 is the only dataset where the solo ranking approach beats ensembles. The reason for this is, on the one hand, a small overlap with relations from the train part and, on the other hand, the high similarity between the question substring and the gold relation. As for Mintaka, here GPT-like based models beat our baseline. Probably, it's connected with the small amount of questions, or Mintaka may have been a part of the training data.

\paragraph{Method efficiency.} Being comparable in terms of accuracy with other approaches, our method is also lightweight. For entity linking, we used the mGENRE approach with 406M parameters \cite{de2020autoregressive} and BELA, which has inside the XLM-R large model with 550M parameters \cite{goyal2021larger}. Relation classification uses the bert-base-uncased model, and relation ranking uses the distilbert-base-uncased model with 110M and 66M of parameters, respectively\footnote{\url{https://huggingface.co/transformers/v2.9.1/pretrained_models.html}}, label generation is based on T5 large model with 770M of parameters~\cite{roberts-etal-2020-much}. In contrast, FlanT5-xl \cite{chung2022scaling}  and t5-xl-ssm-nq \cite{2020t5} have about 3B parameters; GPT-3 –- 175B parameters \cite{ye2023comprehensive}. In addition, other approaches such as KEQA and M3M use Pytorch-BigGraph embeddings, which result in an additional 40GB of memory usage.\footnote{\url{https://torchbiggraph.readthedocs.io/en/latest/pretrained_embeddings.html}}
\vspace{-1ex}
\paragraph{Additional results.}

\begin{figure*}[ht!]
\begin{center}
\includegraphics[width=\textwidth, height = 8cm]{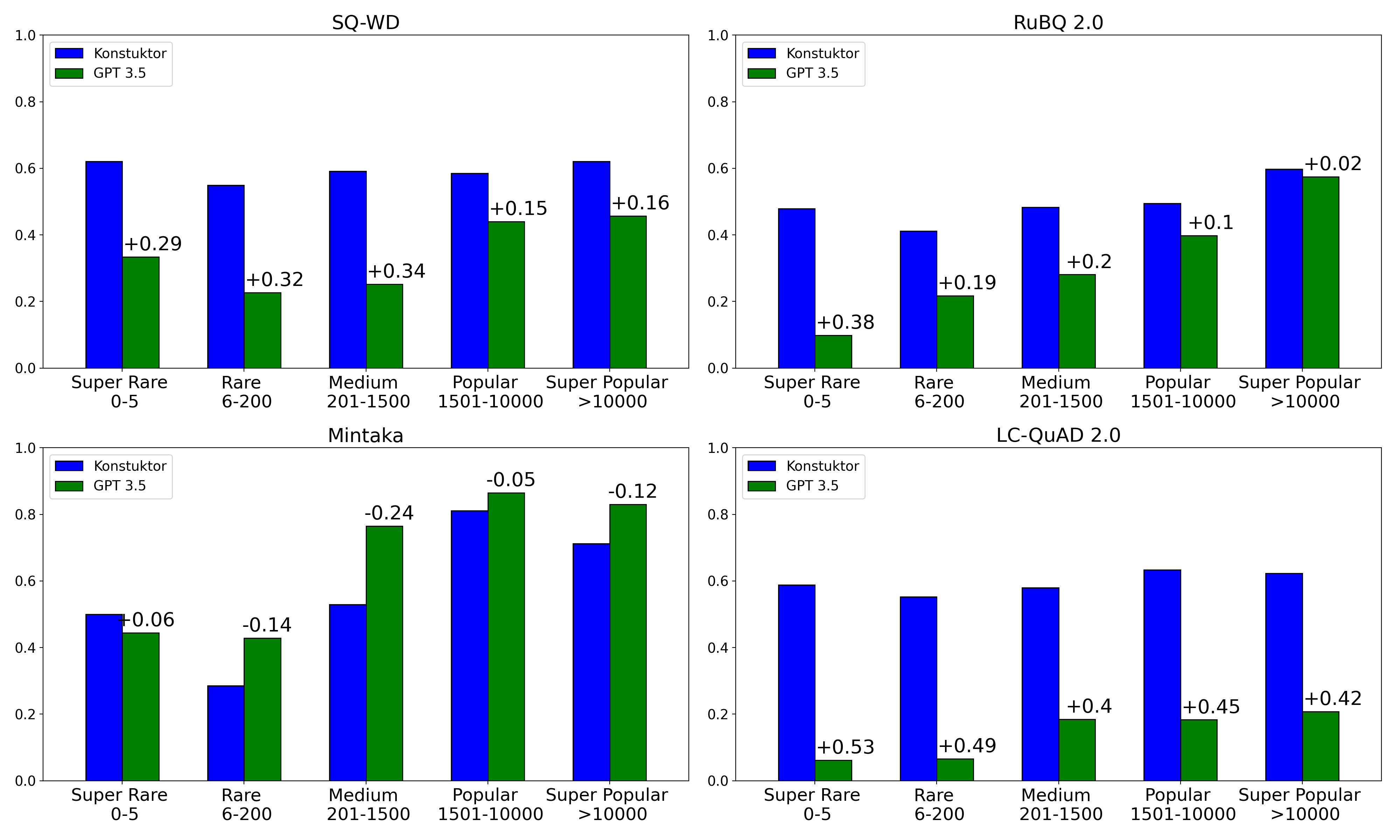} 
\caption{Accuracy depending on the minimum popularity of the question and answer entities. Popularity is measured as corresponding Wikipedia pageviews.
}
\label{results_by_popularity}
\end{center}
\end{figure*}

To further strengthen our results and prove the necessity of developing non-generative QA methods, we split accuracy results obtained in Table~\ref{tab:results} by popularity of the entities from the question and from the answer. The popularity is measured as an amount of page views of the respected Wikipedia page. The data is collected via Wikimedia REST API.\footnote{\url{https://wikimedia.org/api/rest\_v1/}}. The results are presented in Figure~\ref{results_by_popularity}. For each popularity range, the difference in accuracy between Konstruktor's approach and GPT 3.5 is calculated. There are two important observations. First, our approach on average answers questions with rare entities considerably better than GPT 3.5. Second, for the RuBQ dataset we see the vivid trend: the more popular is the entity, the less is the difference in obtained accuracy. Both observations prove that generative approaches fall behind answering even simple questions for rare entities and development of the alternative methods such as \texttt{Konstuktor} is needed.

\section{Ablation Study}

\label{Ablation}







\subsection{Entity Linking and NER}

\label{ent_lin_discussion}
There are two main data pre-processing steps that can potentially improve EL accuracy. The first is the insertion of NER into the sentence. This step is incorporated in the BELA method by default, while the insertion of NER into the mGENRE improves it EL quality, since it has been initially trained  for the entity disambiguation task. This is an example of a sentence without NER insertion: ``Where did Leo Tolstoy die?''. The same sentence with NER insertion is:  ``Where did \textsf{[START]} Leo Tolstoy \textsf{[END]} die?''

Secondly, there is an observation that EL systems may be sensitive to the capitalisation of the words in question \cite{sakor2019old}. An example of a question with NER insertion and uppercase: ``Where Did \textsf{[START]} Leo Tolstoy \textsf{[END]} Die?''



\vspace{-3ex}

\hspace{-5cm}
\begin{table}[htb!]
\footnotesize
\centering

\renewcommand{\arraystretch}{1.1}
\caption{mGENRE and BELA entity linking results in terms of Accuracy@1 for the SQ-WD and Mintaka datasets (validation); RuBQ (test). The best scores are highlighted. }
\scalebox{0.81}{
\hspace{-0.3cm}
\begin{tabular}{lcccccccc}
\toprule
               & \multicolumn{6}{c}{{ \textbf{mGENRE}}}                                         & \multicolumn{2}{c}{{ \textbf{BELA}}}                                                         \\ \toprule
               & \begin{tabular}[c]{@{}c@{}}no NER\\ InitCase\end{tabular} & \begin{tabular}[c]{@{}c@{}}no NER\\ UpperCase\end{tabular} & \begin{tabular}[c]{@{}c@{}}Spacy NER\\ InitCase\end{tabular} & \begin{tabular}[c]{@{}c@{}} Spacy NER\\ UpperCase\end{tabular}& \begin{tabular}[c]{@{}c@{}}BELA NER\\ InitCase\end{tabular} & \begin{tabular}[c]{@{}c@{}}BELA NER\\ UpperCase\end{tabular}& \begin{tabular}[c]{@{}c@{}}BELA NER\\ InitCase\end{tabular} & \begin{tabular}[c]{@{}c@{}}BELA NER\\ UpperCase\end{tabular}  \\ \midrule
SQ-WD & 0.525	& 0.729	&	0.742	& \textbf{0.874}	& 0.615 &	0.724 &	0.702 &	0.779 \\
RuBQ* &  -- &  -- &  0.709  & 0.679 & 0.783 & 0.769 &  \textbf{0.804}  & 0.771\\
Mintaka*&  -- &  -- &  0.883 & 0.865 & 0.820  & 0.813  &  \textbf{0.897} & 0.865\\
\bottomrule
\end{tabular}

}
\label{ent_lin_short}
\end{table}

\vspace{-2ex}

For the mGENRE model, we use two NERs. The first one, Spacy, is fine-tuned on the train parts of the respective datasets.  Second, with the mGENRE model, we try the NER obtained by the BELA model. Using two NER models and InitCase\slash Uppercase variations, there are 4 variants for the input to the mGENRE model.

For the BELA model, there is an internal NER step, and there is no need for a substitution for other NER systems.  Therefore, there are only two input variants, including the InitCase and Uppercase variants.





\paragraph{Results} for the EL stage  are presented in Table~\ref{ent_lin_short} for both mGENRE and BELA models. For the SQ-WD dataset, the best configuration is with Spacy NER, mGENRE model, and upper case, while for other datasets, the BELA model with the initial case prevails. The initial case works badly for the SQ-WD since in this dataset, real named entities are mostly often not capitalized, which is generally the case in other datasets.


\begin{table}[ht!]
\small
\centering
\setlength\tabcolsep{0.35cm}
\renewcommand{\arraystretch}{1}
\caption{Comparison of low case letters ratios for all datasets: SQ-WD (validation), RuBQ-en* (test), Mintaka*(validation). Best scores are highlighted. *Subset of 1-hop questions.}
\scalebox{0.96}{
\begin{tabular}{lccc}

\toprule
        & \begin{tabular}[c]{@{}c@{}} \bf SQ-WD \end{tabular} & \begin{tabular}[c]{@{}c@{}} \bf RuBQ-en* \end{tabular} & \begin{tabular}[c]{@{}c@{}} \bf Mintaka* \end{tabular} \\ \midrule
        
\multicolumn{1}{c}{\begin{tabular}[c]{@{}c@{}}Ratio of subjects in  low case\end{tabular}} & 0.975                              & 0.135                             & 0.141                                \\
BELA NER score                                                                               & 0.674                              & \textbf{0.708}                             & \textbf{0.839}                                  \\
Spacy NER score                                                                              & \textbf{0.898}                              & 0.543                             & 0.747                               \\ \bottomrule
\end{tabular}
}
\label{low_case}
\end{table}

\paragraph{Discussion.} 

In Table ~\ref{low_case}, it is shown that SQ-WD has a disproportionately large number of subjects in low case contrast to other datasets. This influences the scores of the NER models, which in turn influences the scores of the EL models. The low score of the pre-trained BELA model for both NER and EL can be explained by the fact that it was trained on more "natural" data (named entities in most cases contain capitalized letters). That's why, for the SQ-WD dataset, the combination of capitalisation and custom-trained NER significantly boosted performance.
This observation may serve as a guideline for choosing EL system.



\subsection{Relation Detection}

\paragraph{Description.} In Table~\ref{rel_clf} are presented results of the relation detection on the test of the respective datasets. The results are presented for all datasets except Mintaka since it has no labelled relations.

In the first six rows are presented results for the relation classification that differ by the data they were trained on and the loss used in training. SQ-WD data means the training and validation part of the SQ-WD dataset.  All data includes train and validation parts of the SQ-WD dataset, the development part of the RuBQ-en* dataset, 380 questions from the Mintaka train part labelled semi-automatically and SMART dataset. All data with an asterisk means that "direct" and "reversed" relations (described in the Section \nameref{sparql_queries}) were treated the same (in the case of the P19 and R19 relations they were labelled as \textit{place of birth} in both cases). BCE loss is a standard Binary Cross Entropy loss while DB loss is a special Distribution-Balanced loss.

\vspace{-5pt}

\begin{table}[ht!]
\small
\centering
\renewcommand{\arraystretch}{1.1}
\caption{Relation detection results in terms of Accuracy@1 for the SQ-WD (test), RuBQ 2.0*(test), and LC-QuAD 2.0* datasets. The best scores are highlighted. *Subset of 1-hop questions. **Best score on inference out of all existing models.}
\setlength\tabcolsep{0.3cm}
\scalebox{0.96}{
\begin{tabular}{lccc}

\toprule
\bf Method/Train data/Loss                                                                                 & \bf SQ-WD          & \bf RuBQ-en* & \bf LC-QuAD 2.0*      \\ \midrule
Classification: SQ-WD+BCE loss &	0.940	&0.299 & 0.019 \\
Classification: SQ-WD+DB loss	& \textbf{0.949} &	0.353 & 0.019 \\
Classification: all data+BCE loss	& 0.940	& 0.379 &0.033 \\
Classification: all data+DB loss	& \textbf{0.949}	& 0.412  & 0.034\\
Classification: all data*+BCE loss	& 0.719 & 	0.562  & 0.352\\
Classification: all data*+DB loss	& 0.729	& \textbf{0.653}  & 0.360\\
Ranking	& 0.630	& 0.573 & 0.609\\
Label generation	& 0.725	& 0.632  & 0.674 \\
\midrule
Falcon 2.0 \cite{sakor2020falcon}	& 0.278	& 0.174  & 0.568 \\
EARL \cite{dubey2018earl}	& 0.072	& 0.146  & 0.149 \\
GenRL \cite{rossiello2021generative}	& 0.731	& 0.509**  & \textbf{0.805} \\

\bottomrule
\end{tabular}
}
\vspace{-8pt}

\label{rel_clf}
\end{table}

\paragraph{Discussion.} From the Table~\ref{rel_clf}, it may be inferred that the results with DB loss are better in all cases for all datasets. This supports the hypothesis of the applicability of the method to the relation classification task (it was initially tested on the multi-label text classification task). However, for the RuBQ-en* dataset, the discrepancy between the losses is more severe because the distribution of its relations in the training data is not initially balanced, as is the case for the SQ-WD dataset.

Overall, the classification results outperformed the other approaches, with the exception of LC-QuAD 2.0. For LC-QuAD only 34\% of the data was the part of the train. However, the label generation and ranking approaches performed well due to a disproportionate number of cases where the relation label
is the exact substring of the question. According to the results of the ensemble approach, the relation classifier \textit{all data+DB loss} for the SQ-WD dataset and the relation generation for other datasets were chosen as the basic relation detection technique for the final pipeline. 

Comparing \texttt{Konsturktor} with other approaches, Falcon 2.0 \cite{sakor2020falcon} and EARL \cite{dubey2018earl} fall behind our results, while GenRL \cite{rossiello2021generative} shows competitive performance. Our approach beats it on SQ-WD and RuBQ-en datasets. While GenRL~\cite{rossiello2021generative} was trained individually for SQ-WD and LC-QuAD data, for the RuBQ we used all existing models for inference. The worst result of GenRL \cite{rossiello2021generative} is 0.253 obtained by the model trained on the LC-QuAD 1.0 data and 0.509 trained on the LC-QuAD 2.0 data. So, the variance in scores is much higher in comparison to all our relation detection approaches.


\vspace{-2ex}

\subsection{Error Analysis}

In Table~\ref{error_analysis} the results  are provided for the best pipelines of the test set of the respective datasets. For the SQ-WD dataset, the main bottlenecks are the NER and entity linking models, due to the high relation classification accuracy alone.  Vice versa, for other datasets, the main bottleneck is the relation detection module. Although the entity linking results are compatible for all datasets, for the RuBQ-en* dataset, the problem of unseen relations in train is more vivid because of their lower representation in train. 

\vspace{-3ex}

\vspace{-5pt}
\begin{table}[h!]
\small
\centering
\setlength\tabcolsep{0.35cm}
 \caption{Error analysis by components.
*Subset of 1-hop questions.}
\scalebox{0.96}{

\begin{tabular}{lccc}
\toprule
        & \begin{tabular}[c]{@{}c@{}} \bf Entity Linking \end{tabular} & \begin{tabular}[c]{@{}c@{}} \bf Relation Detection \end{tabular} & \begin{tabular}[c]{@{}c@{}}  \bf  Accuracy \end{tabular}  \\ \midrule
SQ-WD                            & 0.773                                                              & 0.949                                                                       & 0.742                                                                         \\
RuBQ-en*                             & 0.712                                                              & 0.632                                                                       & 0.451                                                                       \\
Mintaka*                              & 0.838                                                           & -                                                                           & -                                                                          \\ 
LC-QuAD 2.0                           & 0.840                                                              & 0.609                                                                       & 0.592                                                                       \\
\bottomrule
\end{tabular}
}
\vspace{-10pt}
\label{error_analysis}
\end{table}

\vspace{-5pt}

\begin{table}[ht!]
\small
\centering
\renewcommand{\arraystretch}{1.1}
\caption{Relation detection analysis: accuracy scores corresponding to \textbf{SQ}-WD, \textbf{Ru}BQ, and \textbf{LC}-QuAD datasets. The best scores for the method and dataset are highlighted.
}
\scalebox{0.96}{
 \setlength\tabcolsep{0.17cm}
\begin{tabular}{lccc|ccc|ccc}

\toprule

& \multicolumn{3}{c}{\bf Classification} & \multicolumn{3}{c}{ \bf  Generation} &  \multicolumn{3}{c}{\bf  Ranking} \\ 
& SQ & Ru & LC &  SQ & Ru & LC &  SQ & Ru & LC  \\ \midrule

Not seen (new) relation & -- & 0 & 0.04 & 0   &  0  & 0.41 & 0  & \textbf{0.01}  & \textbf{0.43}    \\
High relation  frequency & \textbf{0.95} & 0.5 & \textbf{0.03}  & 0.72 & \textbf{0.53} & \textbf{0.03} & 0.63 & 0.44 & 0.02 \\
Low relation frequency & 0 & \textbf{0.15} & \textbf{0.29} & 0 & 0.11 & 0.23 & 0 & 0.13 & 0.16 \\ 
\bottomrule
\end{tabular}
}
\vspace{-10pt}

\label{rel_analysis}
\end{table}


Due to a large amount of relation detection techniques, in Table~\ref{rel_analysis} are presented cases to better choose the technique according to data peculiarities. If there is a large amount of relations outside the train, then, ranking procedure is the one to choose. In case of the rich balanced training data both classification and generation techniques are acceptable. Due to the specially constructed DB loss, in case of scarce unbalanced training data, classification with DB loss is the optimal procedure.


\vspace{-1ex}
\section{Conclusion}
In this study a new method \texttt{Konstruktor} for the QA task has been proposed. It is lightweight and open source.\footnote{
\url{https://github.com/s-nlp/konstruktor}} The components include two NER models, two entity linkers and three relation detection techniques.

In our experiments, we showed that a simple KG-based workflow still can outperform modern end-to-end neural models based on LLMs for simple question QA on several datasets. Namely, \texttt{Konstruktor} showed results comparable to SOTA and outperformed approaches such as GPT, on three out of four datasets. Furthermore, we demonstrate  that the benefit of the suggested method is more pronounced when answering questions about rare named entities.
\bibliographystyle{splncs04}
\bibliography{mybibliography}
%




\end{document}